\documentclass[letterpaper]{article} 
\usepackage{aaai2027}  
\usepackage[hyphens]{url}  
\usepackage{graphicx} 
\usepackage{natbib}  
\usepackage{caption} 
\usepackage{algorithm}
\usepackage{algorithmic}
\usepackage{newfloat}
\usepackage{tabularx}
\usepackage{listings}
\usepackage{booktabs}
\usepackage{amsmath}
\usepackage{amssymb}

\usepackage[table]{xcolor}
\usepackage{arydshln}
\usepackage{multirow}

\DeclareCaptionStyle{ruled}{labelfont=normalfont,labelsep=colon,strut=off} 
\floatstyle{ruled}
\newfloat{listing}{tb}{lst}{}
\floatname{listing}{Listing}

\usepackage{booktabs}

\nocopyright

\title{MuST-VAD: Mutual Structured Learning for Video Anomaly Detection}
\author{
Satoshi Hashimoto\textsuperscript{\rm 1}\corresponding,
Hitoshi Nishimura\textsuperscript{\rm 1},
Mori Kurokawa\textsuperscript{\rm 1}
}

\affiliations{
\textsuperscript{\rm 1}KDDI Research, Inc.\\
Fujimino, Saitama, Japan\\
st-hashimoto@kddi.com
}

\begin{document}

\maketitle

\begin{abstract}
In this paper, we propose MuST-VAD, a mutual structured learning framework for weakly supervised video anomaly detection (VAD) in which an anomaly detector and a large vision--language model (LVLM) exchange their acquired knowledge. Detectors in weakly supervised VAD learn anomaly scores from features extracted by a fixed, task-agnostic backbone. These fixed features bound the achievable detection accuracy. Recent methods therefore transfer LVLM semantics into the detector as richer features. However, this transfer is one-way: what the detector learns about the target videos never returns to the LVLM. MuST-VAD extends the one-way transfer into a bidirectional learning loop. In this loop, the latest detector predictions supervise the LVLM adaptation, and the adapted LVLM returns updated representations that retrain the detector; the two models alternate these updates over small video groups. Both models train on detector-selected key clips, while confidence weighting and annotation-anchored question answering keep the exchanged supervision reliable. On UCF-Crime, our mutual learning improves the one-pass transfer baseline from 88.15\% to 88.63\% AUROC and from 37.25\% to 42.46\% average precision (AP), outperforming the state-of-the-art method in AP by 4.13 points.
\end{abstract}

\section{Introduction}
\label{sec:intro}

Video anomaly detection (VAD) aims to temporally localize anomalous events such as fights, road accidents, and thefts in untrimmed surveillance videos. Because annotating the precise temporal extent of every anomaly is prohibitively expensive at deployment scale, weakly supervised VAD (WSVAD) has become the dominant setting: a model is trained with only video-level normal/abnormal labels, yet must produce segment-level anomaly scores at test time. WSVAD is commonly cast as multiple instance learning (MIL), which treats a video as a bag of segments and assumes that an abnormal video contains at least one abnormal segment \citep{sultani2018real}. MIL detectors are attractive in deployment: they are lightweight, train from cheap labels, and years of refinement have made their local discrimination remarkably sharp \citep{tian2021weakly,zhou2023dual}.

However, an MIL detector can only carve its decision boundary in the space spanned by its pre-extracted input features. Real-world anomalies are defined not only by appearance but also by object interactions, action semantics, and scene context, so fixed features from a task-agnostic visual backbone often leave semantically subtle anomalies entangled with normal activity. The detector, no matter how well designed, inherits this ceiling from a feature extractor that knows nothing about the detection task.

\begin{figure}[t]
\centering
\includegraphics[width=0.98\columnwidth]{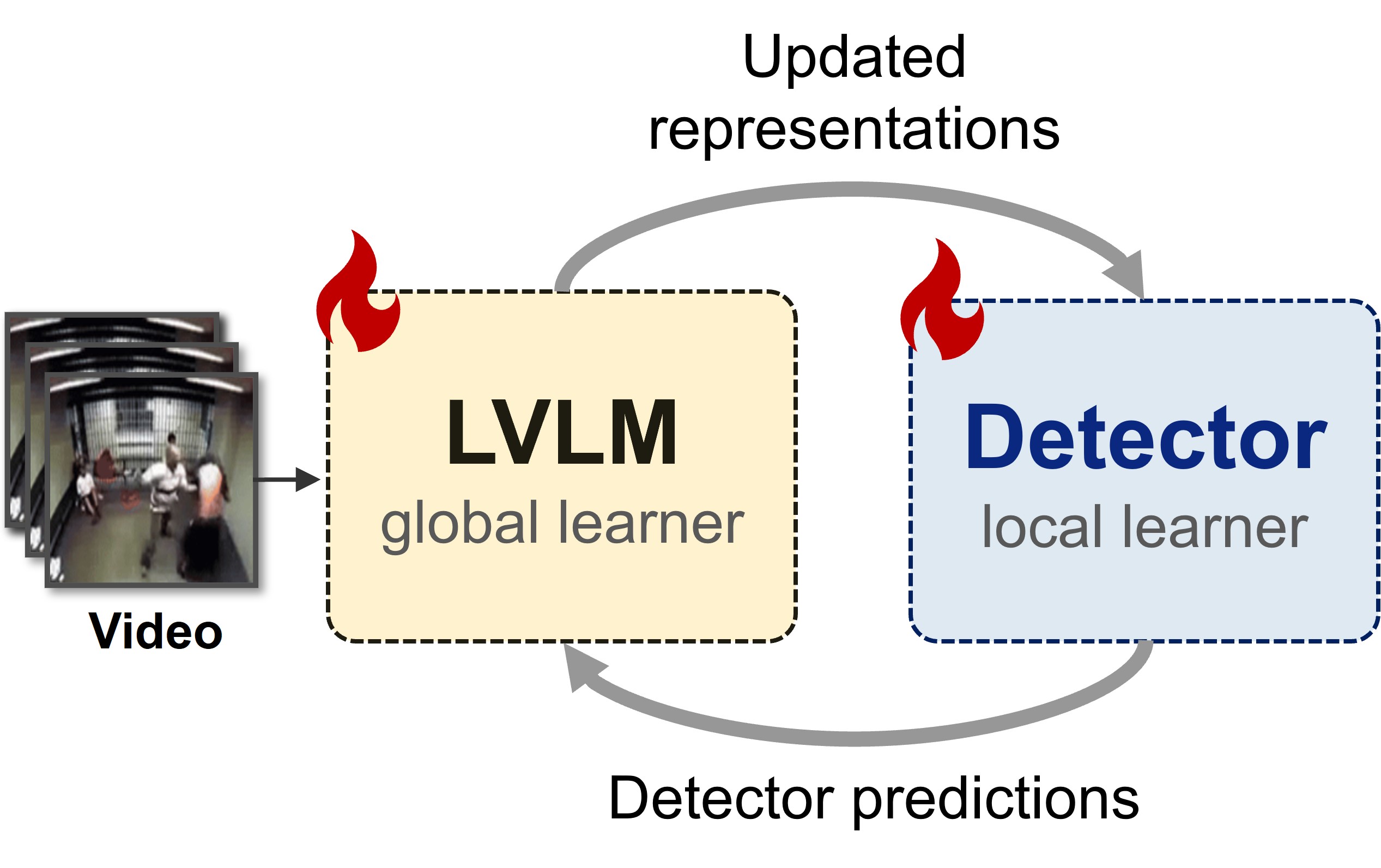}
\caption{Overview of MuST-VAD. A global LVLM and a local detector are trained jointly around a closed loop: the LVLM supplies updated video representations to the detector, and the detector returns predictions that supervise the LVLM.}
\label{fig:overview}
\end{figure}

A growing body of work therefore imports the semantics of pretrained vision-language models into WSVAD, either by aligning anomaly-related text prompts with frozen CLIP representations \citep{wu2024vadclip,yang2024text}, or by employing large vision-language models (LVLMs) as captioners \citep{zanella2024harnessing}, rule inducers \citep{yang2024follow}, and instruction-tuned anomaly assistants \citep{zhang2025holmes}; we review these families in the Related Work section. Such methods confirm that pretrained semantics can lift the feature ceiling. Yet the information flows in one direction, from the pretrained model into the detection pipeline. What the local detector learns about the target domain---which segments are anomalous, when they occur, and how strongly---never returns to the semantic model as a learning signal. 
We propose \textbf{MuST-VAD}, a mutual structured learning framework that closes this loop through an interaction between an LVLM and an anomaly detector (Figure~\ref{fig:overview}). The LVLM provides semantically enriched segment representations to the detector, while the detector returns prediction-derived supervision that adapts the LVLM to the target videos. Structured question-answering tasks derived from video-level annotations provide stable supervision for LVLM adaptation. The two models are alternately updated over small bag groups, allowing newly acquired knowledge to be exchanged throughout training, while confidence-aware supervision reduces the influence of unreliable detector predictions. Through this iterative process, the semantic representations and the anomaly detector are progressively refined together. Our contributions are threefold.
\begin{itemize}
    \item We introduce MuST-VAD, a mutual structured learning framework that closes the iterative learning loop between an LVLM and a weakly supervised anomaly detector, enabling each model to benefit from knowledge acquired by the other.

    \item We develop a mutual learning strategy that combines structured QA supervision, detector-guided key-clip selection, alternating updates over small bag groups, and confidence-aware feedback to enable knowledge exchange between the two models.

    \item We evaluate the complete loop and its components on UCF-Crime under a controlled protocol that uses the same LVLM, detector, and inference procedure as one-pass representation transfer. One mutual round improves both metrics over this baseline and raises frame-level AP to $42.46\%$, surpassing the best previously reported value by $4.13$ points.
\end{itemize}

\section{Related Work}
\label{sec:related}

\subsection{Weakly Supervised Video Anomaly Detection}
\label{sec:wsvad-formulation}

Weakly supervised video anomaly detection localizes anomalous temporal segments while using only video-level normal/abnormal labels during training. \citet{sultani2018real} introduced the canonical MIL formulation for real-world surveillance videos: each untrimmed video is treated as a bag of temporal segments, every segment in a normal bag is assumed to be normal, and an abnormal bag is assumed to contain at least one anomalous segment. Their detector is optimized to rank the highest-scoring segment in an abnormal bag above the highest-scoring segment in a normal bag. We summarize this formulation below because it underlies many subsequent WSVAD methods.

Let $\mathcal{D}=\{(V_i,y_i)\}_{i=1}^{N}$ denote a weakly supervised VAD training set, where $V_i$ is an untrimmed video and $y_i\in\{0,1\}$ is its video-level label, with $y_i=1$ indicating an abnormal video. Each video is divided into temporal segments and represented as a bag of instances. We denote a normal bag and an abnormal bag by
\begin{equation}
\widetilde{B}_{n}=\{\mathbf{f}_{n}^{t}\}_{t=1}^{T_n},
\qquad
\widetilde{B}_{a}=\{\mathbf{f}_{a}^{t}\}_{t=1}^{T_a},
\label{eq:bags}
\end{equation}
where $\mathbf{f}^{t}\in\mathbb{R}^{d}$ is the $d$-dimensional feature of the $t$-th segment. A detector $D_{\phi}$ assigns an anomaly score to each segment,
\begin{equation}
s^{t}=D_{\phi}(\mathbf{f}^{t}).
\label{eq:detector}
\end{equation}
Under the MIL assumption, every segment in a normal bag is normal, whereas an abnormal bag contains at least one anomalous segment. The canonical ranking objective introduced by \citet{sultani2018real} is
\begin{equation}
\mathcal{L}_{\mathrm{MIL}}
=
\max\!\left(
0,
1-\max_{1\leq t\leq T_a}s_{a}^{t}
+\max_{1\leq t\leq T_n}s_{n}^{t}
\right).
\label{eq:mil_rank}
\end{equation}
This loss is zero when the largest anomaly score in an abnormal bag exceeds the largest score in a normal bag by at least one; otherwise, it penalizes the violated margin. It therefore converts video-level supervision into a learning signal for segment-level anomaly scores without requiring temporal annotations.

Subsequent work improves this basic formulation in two main directions. One line strengthens local discrimination inside the weakly supervised objective: RTFM learns robust temporal feature magnitudes \citep{tian2021weakly}, MGFN contrasts feature magnitudes across scenes with a glance-and-focus mechanism \citep{chen2023mgfn}, CMRL exploits relations between context and motion \citep{cho2023look}, and UR-DMU separates normal and abnormal prototypes with dual memory units under uncertainty regulation \citep{zhou2023dual}. Another line refines supervision by self-training: MIST converts MIL outputs into segment-level pseudo labels that fine-tune a task-specific encoder \citep{feng2021mist}, CU-Net models the completeness and uncertainty of pseudo labels \citep{zhang2023exploiting}, and UMIL partitions segments into confident and ambiguous sets to reduce contextual bias \citep{lv2023unbiased}. These methods retain the same weak-label setting while replacing or augmenting the canonical ranking loss with richer detector-specific objectives. MuST-VAD preserves the native objective of the selected detector; it changes how the detector's input representation evolves and adds auxiliary bidirectional distillation.

\subsection{VLM-Adapted Anomaly Detection}

A second family transfers vision-language semantics into WSVAD. Early work uses CLIP as a stronger feature extractor with temporal self-attention \citep{joo2023clip}. Subsequent methods adapt the language interface itself: VadCLIP attaches to a frozen CLIP a dual branch for coarse binary classification and fine-grained text alignment \citep{wu2024vadclip}, AnomalyCLIP identifies a normal subspace and text-driven anomaly directions in the CLIP latent space \citep{zanella2024delving}, and TPWNG generates frame-level pseudo labels from image-text matching under normality guidance and self-trains the classifier \citep{yang2024text}. PEL4VAD incorporates external commonsense knowledge into textual prompts \citep{pu2024learning}, whereas PE-MIL learns abnormal-aware and normal-context prompts for MIL-based anomaly discrimination \citep{chen2024prompt}. Beyond binary detection, OVVAD targets open-vocabulary anomalies with LLM-derived semantic knowledge \citep{wu2024open}, Fine-VAD aligns binary, macro-category, and fine-category supervision progressively \citep{zhang2026fine}, and DSANet disentangles event and background semantics under normality alignment \citep{dsanet2026}. Most CLIP-based methods above optimize prompts, alignments, or pseudo labels within a single VLM-centered detection pipeline, without repeatedly updating a separate LVLM representation learner and MIL detector. PEL4VAD additionally combines pretrained I3D features with representations from a CLIP text encoder.

\subsection{LVLM-Based Anomaly Understanding}

A third family employs LVLMs whose native output is language. Training-free systems score anomalies by aggregating frame captions with a language model \citep{zanella2024harnessing}, by inducing normality rules from a few normal frames and applying them deductively \citep{yang2024follow}, or by verbalized optimization of the guiding questions posed to a frozen LVLM \citep{ye2025vera}. Tuned systems adapt the LVLM itself: VAD-LLaMA equips it with a long-term context module and fine-tunes it in stages \citep{lv2024video}, HAWK learns open-world anomaly description and question answering with an explicit motion modality \citep{tang2024hawk}, and Holmes-VAD and Holmes-VAU instruction-tune a multimodal LLM on frames selected by an anomaly scorer, the latter across temporal granularities with hierarchical instruction data \citep{zhang2024holmesvad,zhang2025holmes}. CUVA benchmarks the causal understanding of video anomalies \citep{du2024uncovering}, and SteerVAD steers anomaly-relevant internal representations of a frozen LVLM \citep{steervad2026}. Where a detector appears, it selects what the LVLM sees; the information flow terminates at the LVLM, whose updated representations are not returned to retrain an independent detector.

\subsection{Iterative and Mutual Learning}

Iterative refinement is well established within a single model: MIST, CU-Net, and TPWNG all alternate between pseudo-label generation and model updates \citep{feng2021mist,zhang2023exploiting,yang2024text}. Across two models, co-training exchanges pseudo labels between views of the same task \citep{blum1998combining}, while deep mutual learning symmetrically aligns the output distributions of homogeneous peer classifiers \citep{zhang2018deep}.

In video anomaly detection, \citet{zaheer2022generative} proposed Generative Cooperative Learning (GCL), which alternately trains an autoencoder generator and a discriminator from entirely unlabeled videos. Reconstruction-error-based pseudo-labels from the generator supervise the discriminator; discriminator-derived pseudo-labels then update the generator through negative learning, which discourages accurate reconstruction of likely anomalies. GCL is therefore an important VAD-specific precedent for iterative cross-supervision between heterogeneous models. MuST-VAD differs in both the supervision and the exchanged information: it retains video-level WSVAD labels and annotation-anchored QA, couples an LVLM representation learner with a MIL detector, and exchanges confidence-weighted segment probabilities and updated semantic representations, not thresholded pseudo-labels between a generator and a discriminator.

Closest to us, caption-based complementary learning couples a detector and an LVLM in both directions---detector scores select key clips and enrich automatically generated captions, and the fine-tuned LVLM returns intermediate features to a detector---but performs this exchange only once \citep{hashimoto2025surveillance}. In summary, prior work explores detector self-training, cooperative learning between generative and discriminative anomaly models, vision-language prompt optimization, score-guided selection of LVLM inputs, and one-pass complementary use of a detector and an LVLM. To the best of our knowledge, however, no prior WSVAD method repeatedly uses confidence-weighted predictions from a separate MIL detector to adapt an LVLM representation learner and then retrains that detector on the updated LVLM representations.

\section{MuST-VAD}
\label{sec:method}

\subsection{Overview}
\label{sec:method-overview}

MuST-VAD addresses a limitation of one-way representation transfer: once an anomaly detector is trained, the segment-level knowledge it acquires is not used to improve the LVLM that produced its input features. Our key idea is to alternate the two models through two asymmetric interfaces. The LVLM transfers semantic segment representations to the detector, while the detector returns local anomaly probabilities and informative temporal evidence to the LVLM. Each model retains its own primary objective, and no gradient is propagated across the interface.

The loop has three supporting mechanisms. First, detector-guided key-clip selection concentrates the updates of both models on informative temporal evidence. Second, structured question answering (QA) combines stable annotation-derived targets with detector-derived timing and score targets. Third, confidence-aware distillation suppresses uncertain cross-model predictions. We combine these mechanisms with bag-group asynchronous optimization so that only a small part of the training set is updated at a time. The remainder of this section introduces these components and then summarizes their interaction.

\subsection{Segment Features and Key Clips}
\label{sec:method-feature}

This subsection formalizes the representation path from the LVLM $G_{\theta}$ to the detector $D_{\phi}$: segment features read from $G_{\theta}$ form the detector input, and the resulting detector scores select \emph{key clips}, the compact temporal evidence on which both models are subsequently trained.

\begin{figure*}[t]
    \centering
    \includegraphics[width=\textwidth]{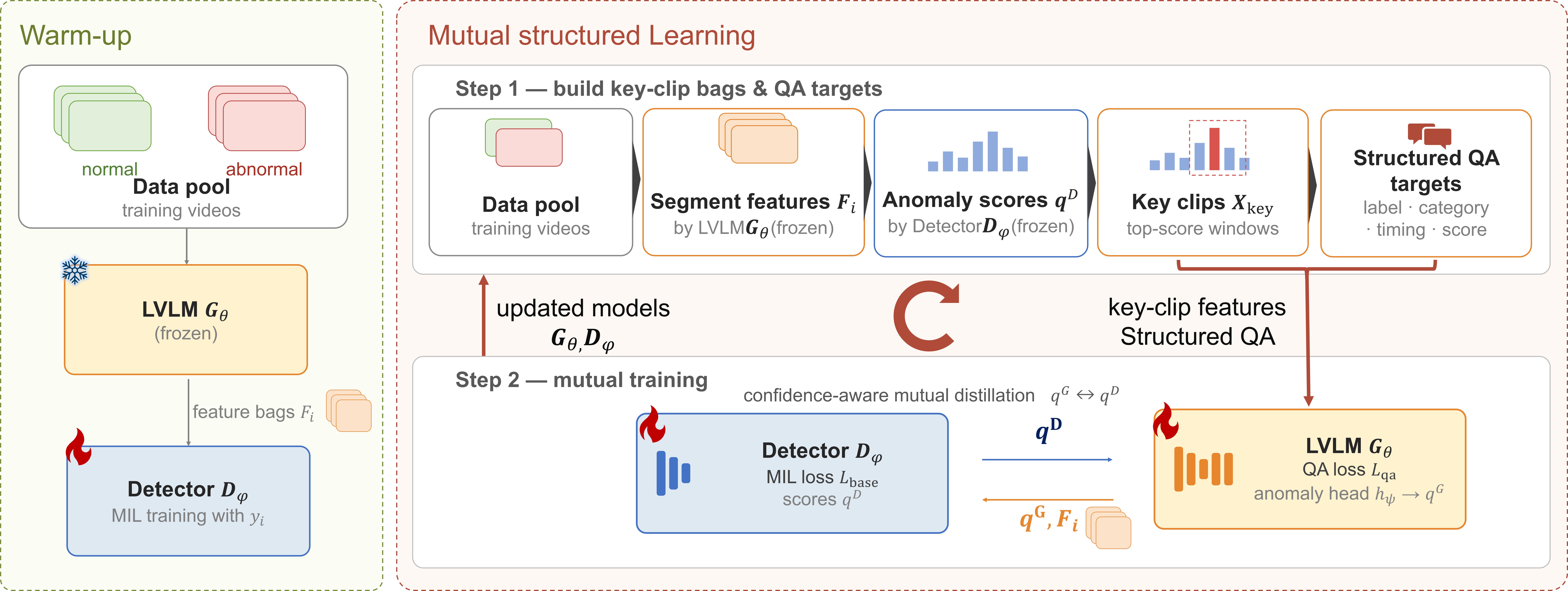}
    \caption{Training workflow of MuST-VAD. During warm-up, the frozen LVLM extracts segment features to train the WSVAD detector. In each mutual-learning cycle, the detector selects key clips and constructs structured QA targets for the LVLM, while the LVLM provides segment representations and anomaly predictions to the detector. The two models are alternately updated through confidence-aware mutual distillation and reused in the next cycle.}
    \label{fig:workflow}
\end{figure*}

Each training video $V_i$ $(i=1,\ldots,N)$ is divided into $T_i$ non-overlapping temporal segments $\{x_{i,t}\}_{t=1}^{T_i}$. To make the contextual semantics of the LVLM available to the detector, we read a fixed-dimensional feature for every segment directly from the hidden states of $G_{\theta}$. Given a segment $x_{i,t}$ and a fixed feature-extraction prompt $q_{\mathrm{feat}}$, let $H_{\theta}^{\mathrm{last}}(x_{i,t},q_{\mathrm{feat}})\in\mathbb{R}^{L_{i,t}\times d}$ denote the hidden states of the final language-model block. With $\mathcal{P}_{i,t}$ denoting the valid multimodal prompt positions, excluding padding, the segment feature is
\begin{equation}
\mathbf{f}_{i,t}(\theta)
=\frac{1}{|\mathcal{P}_{i,t}|}
\sum_{\ell\in\mathcal{P}_{i,t}}
H_{\theta,\ell}^{\mathrm{last}}(x_{i,t},q_{\mathrm{feat}}),
\label{eq:segment_feature}
\end{equation}
and the segment features of $V_i$ are stacked into its complete feature bag $F_i(\theta)\in\mathbb{R}^{T_i\times d}$. The feature is computed in a single prompt-conditioned forward pass and does not depend on any generated text, so yields a fixed-dimensional detector input without text generation.

MuST-VAD is compatible with any WSVAD detector that scores every segment of a feature bag, producing one anomaly logit $z_{i,t}^{D}$ per segment and, through the sigmoid, an anomaly probability $q_{i,t}^{D}$. The detector is trained by minimizing its native weakly supervised objective,
\begin{equation}
\min_{\phi}\;
\mathcal{L}_{D}^{\mathrm{base}}
\!\left(\phi;\{F_i,y_i\}_{i=1}^{N}\right),
\label{eq:base}
\end{equation}
which converts the video-level labels $y_i$ into a learning signal for the segment scores under the MIL assumption of Section~\ref{sec:wsvad-formulation}, as in the canonical ranking loss of Equation~\eqref{eq:mil_rank} or its detector-specific extensions. MuST-VAD leaves the detector architecture and this objective unchanged.

\paragraph{Key-clip selection.}
Untrimmed surveillance videos are dominated by normal segments, so most temporal locations carry little training signal: updating the detector everywhere dilutes hard evidence, and feeding entire videos to the LVLM is computationally prohibitive. We therefore let the current detector decide where both models focus. Ranking the segments of a video by $q_{i,t}^{D}$, we select up to $K_D$ temporally separated high-score segments and take a window of $w_D$ segments around each; we call these windows the key clips of the video. For an abnormal video, the key clips are likely anomaly candidates; for a normal video, the same rule yields hard negatives. Both models are then trained on this shared evidence: the detector on the key-clip features taken from $F_i$, and the LVLM on the raw frames of the top-ranked key clip,
\begin{equation}
t_{i}^{\mathrm{key}}
=\arg\max_{1\leq t\leq T_i}\,q_{i,t}^{D},
\qquad
X_{i}^{\mathrm{key}}
=V_i\!\left[\mathcal{W}\!\left(t_{i}^{\mathrm{key}}\right)\right],
\label{eq:keyclip}
\end{equation}
where $\mathcal{W}(t)$ is a temporal window in which the selected segment $t$ is placed at a randomly sampled position, and $V_i[\cdot]$ extracts the corresponding raw frames. Both selection and placement are discrete, and no gradient passes through either operation.

\subsection{Structured QA}
\label{sec:method-qa}

This subsection describes the reverse path of the loop: how the detector's knowledge supervises the LVLM. Structured QA is the primary objective for adapting $G_{\theta}$ to the target videos. Video-level annotations are stable but contain little temporal information, whereas detector predictions provide temporal detail but evolve during training; Structured QA combines sources so that the LVLM learns from detector feedback without losing reliable weak supervision. All questions are answered from the key clip $X_i^{\mathrm{key}}$ defined above.

The task set contains four questions. Two annotation-derived tasks predict the normal/abnormal label and the anomaly category. Two detector-derived tasks describe the selected temporal evidence. The \emph{detect timing} task identifies the position of the detector-selected segment within $X_i^{\mathrm{key}}$. The \emph{detect score} task predicts the quantized level of the detector score assigned to that segment. The annotation-derived answers remain fixed, while the detector-derived answers are regenerated from the latest detector scores. Thus, the former anchor the video semantics and the latter communicate where the detector attends and how strongly it scores the observed evidence. The task set is configurable, and an experiment may activate any subset of the four tasks.

For the set $\mathcal{B}_{G}$ of videos in one LVLM update, let $\mathcal{K}_i$ be the active tasks for video $V_i$ and $a_{i,k}$ the answer for task $k$. We denote the effective weight of the corresponding QA pair by $\omega_{i,k}$. For annotation-derived tasks, this weight is a fixed per-task constant $\lambda_k$. For detector-derived tasks, the task constant is further multiplied by the confidence of the detector target, and a target below a confidence threshold is omitted. The QA objective is
\begin{equation}
\mathcal{L}_{G}^{\mathrm{qa}}
=-\sum_{i\in\mathcal{B}_{G}}
\sum_{k\in\mathcal{K}_i}
\omega_{i,k}
\log P_{\theta}
\!\left(a_{i,k}\mid X_i^{\mathrm{key}},\mathcal{C}_{i,k}\right),
\label{eq:vlmloss}
\end{equation}
where $P_{\theta}$ is the token probability of the LVLM and $\mathcal{C}_{i,k}$ is the conversation context of task $k$, namely the preceding QA turns and the current question. The active tasks are serialized as a multi-turn conversation, and only assistant answer tokens contribute to the loss. Thus, structured QA provides stable supervision using the detector's temporal knowledge.

\subsection{Confidence-Aware Bidirectional Distillation}
\label{sec:method-distill}

Structured QA transfers a small set of interpretable attributes, but it does not expose the full segment probability sequence to the other model. We therefore couple the two models with an auxiliary distillation loss in both directions, while each model keeps its own primary objective. Predictions and updated representations are exchanged without cross-model gradients, preserving separate model optimization paths.

A lightweight anomaly head $h_{\psi}$ maps each LVLM segment feature to an anomaly logit $z_{i,t}^{G}$ and, through the sigmoid, a probability $q_{i,t}^{G}$, giving the LVLM its own segment-level prediction. Both directions use the same confidence-weighted binary cross-entropy between student logits $z_u$ and detached teacher probabilities $q_u$, where $u$ indexes the valid video--segment pairs $(i,t)$ of the update:
\begin{equation}
\begin{aligned}
\ell(z;q)&=
\frac{\sum_{u} m_u w_u\,\mathrm{BCE}\!\left(z_u,\mathrm{sg}[q_u]\right)}
{\sum_{u}m_u w_u+\varepsilon},\\
w_u&=|2q_u-1|^{\gamma},
\end{aligned}
\label{eq:distill}
\end{equation}
where $\mathrm{sg}[\cdot]$ denotes stop-gradient and $\varepsilon$ is a small constant. The weight $w_u$ emphasizes confident targets, and the binary mask $m_u$ keeps only targets with $q_u\geq\tau_{+}$ or $q_u\leq\tau_{-}$, where $\tau_{-}<0.5<\tau_{+}$. For a video labeled normal, the weak label overrides the teacher: every segment uses target $0$ with the mask and weight set to one.

In the $D\!\rightarrow\!G$ direction, detector probabilities supervise the anomaly head; in the reverse $G\!\rightarrow\!D$ direction, the anomaly head supervises the detector on the key-clip segments. The two auxiliary losses are
\begin{equation}
\begin{aligned}
\mathcal{L}_{G}^{\mathrm{dist}}&=\beta_{DG}\,\ell(z^{G};q^{D}),\\
\mathcal{L}_{D}^{\mathrm{dist}}&=\beta_{GD}\,\ell(z^{D};q^{G}),
\end{aligned}
\label{eq:dist_pair}
\end{equation}
where $\beta_{DG}$ and $\beta_{GD}$ are fixed weights. The former updates $h_{\psi}$ and the trainable LVLM parameters through features recomputed with gradients enabled, while $D_{\phi}$ stays fixed; the latter updates only $D_{\phi}$. Because $h_{\psi}$ starts untrained, $G\!\rightarrow\!D$ is enabled only after an initial delay, whereas $D\!\rightarrow\!G$ is active throughout mutual learning.

\subsection{Bag-Group Asynchronous Optimization}
\label{sec:method-schedule}

Updating both models over the full training set after every exchange would be prohibitively expensive and would delay the use of newly acquired knowledge. MuST-VAD therefore organizes training into \emph{rounds}. Round $0$ is a one-pass warmup: complete feature bags are extracted once from the pretrained LVLM, and the detector is trained on them with $\mathcal{L}_{D}^{\mathrm{base}}$. This round supplies the initial features and the first reliable segment scores, and it coincides with the one-pass transfer baseline of our experiments. Each subsequent mutual round partitions the training videos into small bag groups, each containing a few normal and a few abnormal bags, and the two models are alternately updated group by group in a sequence of \emph{macro-cycles} until every group has been visited.

One macro-cycle updates the detector and the LVLM on distinct bag groups with the objectives
\begin{equation}
\mathcal{L}_{D}=\mathcal{L}_{D}^{\mathrm{base}}+\mathcal{L}_{D}^{\mathrm{dist}},
\qquad
\mathcal{L}_{G}=\mathcal{L}_{G}^{\mathrm{qa}}+\mathcal{L}_{G}^{\mathrm{dist}},
\label{eq:objectives}
\end{equation}
where $\mathcal{L}_{D}^{\mathrm{dist}}$ is included only after the first $W$ macro-cycles of the round. In the detector phase, $D_{\phi}$ scores the complete bags of its groups, selects their key clips, and is updated on the key-clip features with $\mathcal{L}_{D}$. In the LVLM phase, the updated detector scores the bags of the LVLM group, selects their key clips, and generates the detector-derived QA answers and distillation targets. The LVLM and its anomaly head are then updated on these key clips with $\mathcal{L}_{G}$.

After the LVLM update, the complete bags of the LVLM group are re-extracted with the updated LVLM and replace the previous ones. Each stored bag thus reflects the most recent LVLM state that processed its video, and the bags of videos not yet revisited remain temporarily stale until a later macro-cycle covers them. This bounded staleness avoids a full-dataset extraction after every LVLM update. Algorithm~\ref{alg:mustvad} summarizes one mutual round.

\begin{algorithm}[tb]
\caption{One mutual round of MuST-VAD}
\label{alg:mustvad}
\textbf{Input}: one-pass initialized $G_{\theta}$ and $D_{\phi}$; complete feature bags of all training videos; delay $W$\\
\textbf{Output}: updated \(G_{\theta}\) and \(D_{\phi}\)
\begin{algorithmic}[1]
\STATE initialize anomaly head $h_{\psi}$; partition videos into bag groups
\FOR{macro-cycle $c=0,1,\ldots$ until all groups are visited}
\STATE assign distinct bag groups to the detector and the LVLM
\STATE score the detector groups, select key clips, and update $\phi$ with $\mathcal{L}_{D}$ ($\mathcal{L}_{D}^{\mathrm{dist}}$ only if $c\geq W$)
\STATE score the LVLM group with the updated $D_{\phi}$; select key clips; build QA answers and targets $q^{D}$
\STATE update $(\theta,\psi)$ with $\mathcal{L}_{G}$
\STATE re-extract the complete bags of the LVLM group
\ENDFOR

\RETURN \(G_{\theta},D_{\phi}\)
\end{algorithmic}
\end{algorithm}

\paragraph{Inference.}
The mutual-learning loop, key-clip selection, and auxiliary anomaly head are used only during training. At inference, the final LVLM extracts the complete feature bag of each video once, and $D_{\phi}$ predicts anomaly scores for all segments. The inference procedure is therefore identical to one-pass representation transfer.

\section{Experiments}
\label{sec:experiments}

\subsection{Experimental Setup}
\label{sec:exp-setup}

\paragraph{Dataset and metrics.}
We evaluate MuST-VAD on UCF-Crime, a large-scale surveillance benchmark containing thirteen anomaly categories \citep{sultani2018real}. We use the standard weakly supervised split and derive the class label $c_i$ from each abnormal video's filename; normal videos are assigned \textit{Normal}. We report frame-level area under the receiver operating characteristic curve (AUROC) and frame-level average precision (AP). AUROC remains the dominant UCF-Crime metric, while AP complements it under the severe imbalance between normal and anomalous frames, for which a high AUROC can mask poor precision--recall behavior \citep{acharya2026road}. Segment scores are expanded to their corresponding frames, and both metrics are computed over the complete standard test set.

\paragraph{Implementation details.}
For experiments, we instantiate the LVLM with Qwen3-VL-8B-Instruct \citep{qwen3vl} and the detector with the official UR-DMU architecture and objective \citep{zhou2023dual}. Videos are divided into non-overlapping 16-frame segments. Segment features are extracted with the fixed prompt ``\textit{Is this surveillance video normal or abnormal?}'' using the final-block prompt mean in Equation~\eqref{eq:segment_feature}, yielding $d=4096$. For structured QA, the key clip is a 96-frame window (six 16-frame segments) centered at the detector-selected segment. The vision encoder is frozen; the visual projector and final language-model block are optimized with a base learning rate of $1\times10^{-5}$ and a per-round decay factor of $0.7$. The QA weights are $\lambda_{\mathrm{bin}}=1.0$ and $\lambda_{\mathrm{class}}=0.7$.

Round $0$ trains UR-DMU for 25 epochs with learning rate $5\times10^{-5}$ on the initial feature bags. During mutual learning, each bag group contains two normal and two abnormal videos. One macro-cycle updates one detector group and one LVLM group, with one optimizer step per group. We set $\beta_{DG}=0.2$, $\beta_{GD}=0.1$, $\gamma=2.0$, $\tau_{+}=0.8$, and $\tau_{-}=0.2$. Detector-to-LVLM distillation is active from the first macro-cycle, whereas LVLM-to-detector distillation starts after the first five macro-cycles ($W=5$). For key-clip selection, we set \(K_D=1\) and \(w_D=1\): the detector uses the highest-scoring segment, while the LVLM uses the six-segment window above; normal selections serve as hard negatives. The online detector supplies LVLM feedback, selects all key clips, and is used for final evaluation. We run one mutual-learning round. All experiments use four NVIDIA L40S GPUs and a fixed random seed.

\paragraph{Baseline.}
Our primary baseline is the round-$0$ one-pass transfer described in Section~\ref{sec:method-schedule}. It uses the same initial LVLM features and the same UR-DMU detector as MuST-VAD, but performs no mutual feedback. This comparison isolates the effect of the mutual-learning loop.

\subsection{Comparison with Prior Methods}
\label{sec:exp-sota}


\begin{table}[t]
\centering
\small
\setlength{\tabcolsep}{4pt}
\renewcommand{\arraystretch}{1.06}
\begin{tabularx}{\columnwidth}{>{\raggedright\arraybackslash}Xcc}
\toprule
Method & AUROC & AP \\
\midrule
\multicolumn{3}{c}{\textit{Visual WSVAD}} \\
\midrule
MIL \citep{sultani2018real}
    & 75.41 & 25.03$^{\dagger}$ \\
MIST \citep{feng2021mist}
    & 82.30 & -- \\
RTFM \citep{tian2021weakly}
    & 84.30 & 29.46$^{\dagger}$ \\
UR-DMU \citep{zhou2023dual}
    & 86.97 & 35.48$^{\dagger}$ \\
Road Less Seen \citep{acharya2026road}
    & -- & 38.33 \\
\midrule
\multicolumn{3}{c}{\textit{VLM/LVLM-based VAD}} \\
\midrule
LAVAD \citep{zanella2024harnessing}
    & 80.28 & -- \\
VERA \citep{ye2025vera}
    & 86.55 & -- \\
SteerVAD \citep{steervad2026}
    & 87.15 & -- \\
TPWNG \citep{yang2024text}
    & 87.79 & -- \\
VadCLIP \citep{wu2024vadclip}
    & 88.02 & 33.55$^{\dagger}$ \\
DSANet \citep{dsanet2026}
    & \textbf{89.44} & -- \\

\hdashline
MuST-VAD (Round 0)
    & 88.15 & 37.25 \\
\rowcolor[gray]{0.90}
\textbf{MuST-VAD (Round 1)}
    & 88.63 & \textbf{42.46} \\
\bottomrule
\end{tabularx}
\caption{Frame-level performance (\%) on UCF-Crime.
AP values marked with $^{\dagger}$ are recomputed from publicly
available implementations by \citet{acharya2026road};
``--'' denotes that the result is unavailable.}
\label{tab:sota}
\end{table}

Table~\ref{tab:sota} groups prior methods into visual WSVAD and the broader VLM/LVLM-based family. Because most UCF-Crime methods report only AUROC, the AP column combines originally reported values with those recomputed from public implementations by \citet{acharya2026road}, who motivate AP for the severe frame imbalance of this benchmark. MuST-VAD reaches $88.63\%$ AUROC and $42.46\%$ AP; the AP exceeds the best prior value of $38.33\%$ by $4.13$ points, whereas its AUROC remains below the best listed result, attained by DSANet. Two properties of this result stand out. First, our round-$0$ baseline---the UR-DMU detector trained once on the initial LVLM features---already attains $88.15\%$ AUROC and $37.25\%$ AP, exceeding most listed prior methods with reported AUROC and providing a strong one-pass transfer baseline. Second, one mutual-learning round improves this baseline by $5.21$ AP points and $0.48$ AUROC points. The different magnitudes of these gains are examined in the next subsection without attributing them to a specific internal mechanism.

\subsection{Effect of Bag-Group Mutual Learning}
\label{sec:exp-rounds}


Round $0$ is the one-pass baseline, whereas Round $1$ applies the complete bag-group schedule in Algorithm~\ref{alg:mustvad}; the online detector is used throughout mutual learning and provides the MuST-VAD result reported in Table~\ref{tab:sota}. One mutual round improves AUROC by $0.48$ points and AP by $5.21$ points, a $14.0\%$ relative AP gain.

We hypothesize that the larger AP gain arises from where the loop concentrates its updates. Under the severe frame imbalance of UCF-Crime, AP is especially sensitive to precision in the high-score region, whereas AUROC measures the probability that a randomly selected anomalous frame is scored above a randomly selected normal frame and does not itself depend on class prevalence. MuST-VAD updates both models on detector-selected key clips: the strongest false alarms in normal videos and the leading anomaly candidates in abnormal videos. This design focuses learning on high-scored temporal evidence, while the confidence weight in Equation~\eqref{eq:distill} emphasizes decisive predictions. The larger AP than AUROC improvement is therefore consistent with, but does not establish, a stronger effect on high-score precision than on overall pairwise ranking.

\subsection{Ablation Studies}
\label{sec:exp-ablation}

To examine how performance changes when each component is removed, we conduct one-at-a-time ablations while keeping the round-$0$ initialization, bag-group schedule, and update budget fixed. In Table~\ref{tab:ablation}, \emph{Key-clip} denotes detector-guided key-clip selection, and a cross in this column means that both models are instead trained on a uniformly sampled random window per video. \emph{Distill.} denotes the bidirectional distillation terms, whose removal leaves structured QA as the only coupling. \emph{Conf.\ weight} denotes the confidence weighting; its removal sets $w_u=1$, so all distillation targets contribute equally.

\begin{table}[t]
\centering
\small
\setlength{\tabcolsep}{3.5pt}
\renewcommand{\arraystretch}{1.06}
\begin{tabular}{cccccc}
\toprule
Round & Key-clip & Distill. & Conf. weight & AUROC & AP \\
\midrule
0 & -- & -- & -- & 88.15 & 37.25 \\
\midrule
1 & \checkmark & $\times$   & $\times$   & 88.16 & 41.35 \\
1 & \checkmark & \checkmark & $\times$   & 87.80 & 41.56 \\
1 & $\times$   & \checkmark & \checkmark & 86.54 & 38.38 \\
\midrule
\rowcolor{gray!15}
1 & \checkmark & \checkmark & \checkmark
  & \textbf{88.63} & \textbf{42.46} \\
\bottomrule
\end{tabular}
\caption{Component ablation on UCF-Crime (\%).
Round 0 denotes the one-pass baseline without mutual learning.
A cross under ``Key-clip'' denotes random clip selection; a cross under ``Distill.'' or ``Conf.\ weight'' denotes removing that term.
All Round-1 variants use the same Round-0 initialization and update budget.}
\label{tab:ablation}
\end{table}

As can be seen from Table~\ref{tab:ablation}, the full configuration achieves the best AUROC and AP, while the ablations affect the two metrics differently. Replacing key-clip selection with random windows produces the largest degradation: AUROC falls to $86.54\%$, $1.61$ points below the round-$0$ baseline, and the AP gain over round $0$ shrinks from $5.21$ to $1.13$ points. This pattern suggests that detector-guided selection is important to the observed gain. One possible explanation is that random windows more often contain uninformative segments, weakening the QA and distillation targets and the subsequent LVLM feature updates. The aggregate metrics do not directly show that either model is corrupted, but they are consistent with key-clip selection serving as more than an efficiency device.

Removing distillation preserves most of the AP gain ($41.35\%$) but leaves AUROC near the round-$0$ level ($88.16\%$). Uniform distillation yields a lower AUROC ($87.80\%$) than no distillation, whereas the complete confidence-aware configuration reaches $88.63\%$. These observations suggest that QA on selected clips is associated with much of the AP improvement and that confidence-aware distillation may help overall pairwise ranking. One possible explanation is that uniform weighting gives near-boundary detector and LVLM predictions the same influence as confident targets and may therefore introduce noisier supervision. The results are consistent with confidence weighting improving the usefulness of bidirectional distillation, but they do not identify a unique internal causal mechanism.

\subsection{Performance at Low False-Positive Rates}
\label{sec:low-fpr}

\begin{table}[t]
\centering
\small
\setlength{\tabcolsep}{3pt}
\renewcommand{\arraystretch}{1.08}
\begin{tabularx}{\columnwidth}{
    >{\raggedright\arraybackslash}Xcccc}
\toprule
\multirow{2}{*}{Method}
    & \multirow{2}{*}{AP (\%)}
    & \multicolumn{3}{c}{Recall@FPR} \\
\cmidrule(lr){3-5}
    & & 1\% & 2\% & 3\% \\
\midrule
UR-DMU
    & 35.48$^{\dagger}$ & 0.170 & 0.170 & 0.212 \\
VadCLIP
    & 33.55$^{\dagger}$ & 0.109 & 0.155 & 0.217 \\
The Road Less Seen
    & 38.33 & \textbf{0.173} & 0.263 & \textbf{0.336} \\
\midrule
\rowcolor[gray]{0.90}
MuST-VAD
    & \textbf{42.46}
    & 0.160
    & \textbf{0.266}
    & 0.332 \\
\bottomrule
\end{tabularx}
\caption{Low-FPR performance on UCF-Crime. Recall@FPR denotes
recall under the specified FPR constraint. Competing results are
reported by \citet{acharya2026road}; $^{\dagger}$ denotes their
recomputed AP.}
\label{tab:low-fpr}
\end{table}

To assess performance at practically relevant operating
points, we evaluate Recall@FPR following Acharya et al.
(2026). This metric measures the maximum recall attainable
while constraining the FPR to the specified value.
As shown in Table 3, MuST-VAD achieves the highest recall at FPR=2\% and clearly outperforms
UR-DMU and VadCLIP at both 2\% and 3\%. Compared with
The Road Less Seen, MuST-VAD is slightly better at 2\%
(0.266 vs. 0.263) and comparable at 3\% (0.332 vs. 0.336),
but performs worse at the strictest FPR of 1\% (0.160 vs.
0.173). These results show that the AP improvement is accompanied by competitive anomaly coverage under modest
false-positive budgets, although improving recall remains a limitation.

\section{Conclusion}
\label{sec:conclusion}

We presented MuST-VAD, a mutual learning framework for weakly supervised video anomaly detection. An LVLM provides multimodal segment representations, while a MIL detector selects key clips and returns confidence-weighted predictions for LVLM adaptation. The two models are alternately optimized over distinct bag groups, with feature bags re-extracted only for videos used to update the LVLM. Experiments on UCF-Crime demonstrate improvements over one-pass representation transfer under the same model and inference settings.
MuST-VAD enables a large generalist model and a compact specialist to exchange complementary knowledge without a shared end-to-end computation graph. This principle may also benefit other temporal recognition tasks.

\bibliography{aaai2027}

\end{document}